\documentclass[10pt, a4paper]{article}

\usepackage{lrec}
\usepackage{multibib}

\newcites{languageresource}{Language Resources}
\usepackage{graphicx}
\usepackage{tabularx}
\usepackage{booktabs}
\usepackage{soul}
\usepackage{color}

\usepackage{epstopdf}
\usepackage[latin1]{inputenc}
\usepackage{todonotes}
\usepackage{hyperref}
\usepackage{xstring}
\usepackage{caption} 

\title{From Witch's Shot to Music Making Bones -- Resources for Medical Laymen to Technical Language and Vice Versa\\
}

\name{\hspace*{-3mm}\parbox{\textwidth}{\centering
Laura Seiffe,
Oliver Marten,
Michael Mikhailov,
Sven Schmeier,\\
Sebastian M{\"o}ller, 
Roland Roller\\
}\vspace*{2mm}}

\address{
German Research Center for Artificial Intelligence (DFKI),
 \\Speech and Language Technology Lab, Berlin, Germany \\
    firstname.secondname@dfki.de\\}

\abstract{
Many people share information in social media or forums, like food they eat, sports activities they do or events which have been visited. This also applies to information about a person's health status. Information we share online unveils directly or indirectly information about our lifestyle and health situation and thus provides a valuable data resource. If we can make advantage of that data, applications can be created that enable e.g. the detection of possible risk factors of diseases or adverse drug reactions of medications.   
However, as most people are not medical experts, language used might be more descriptive rather than the precise medical expression as medics do. To detect and use those relevant information, laymen language has to be translated and/or linked to the corresponding medical concept. This work presents baseline data sources in order to address this challenge for German. We introduce a new data set which annotates medical laymen and technical expressions in a patient forum, along with a set of medical synonyms and definitions, and present first baseline results on the data.
\\\newline  }

\begin{document}

\maketitleabstract

\section{Introduction}

Every day people generate and share information online which sheds light on our lifestyle and also to a certain extent to the health situation. Provided information might include data about sports activities, food, alcohol and drug intake, but also indirectly about potential risk factors of diseases or possible adverse drug reactions, see e.g. \newcite{abbar2015} 
or \newcite{weissenbacher2018}. Mining for instance adverse drug reactions has a high relevance for the general public as well as for pharmacological companies. As the level of medication intake is generally increasing all over the world, so does the risk of unwanted side effects \cite{karapetiantz2018}.

In most cases, models to extract health related information from text are trained on large annotated data sets, mainly in English language, and on well formed sentences. Text in social media, forums, but also in emails, can differ in terms of sentence structure, writing style and word usage in comparison to news articles or scientific publications. Thinking particularly of health related information, the language used might be more casual and descriptive rather than the precise medical expression, as most people are not medical experts. This makes it difficult to identify the precise technical expression and to link it against a unique concept in a biomedical ontology, in order to e.g. gather further background knowledge.
This makes it difficult to identify the precise technical expression and to link it to a unique concept in a biomedical ontology, in order to e.g. gather further background knowledge. For instance referring to the title of this work, patients might use laymen expressions such as `Hexenschuss' (\textit{lit.:} `\textit{a witch's shot}', known as `\textit{lumbago}') or `Musizierknochen' (\textit{lit.:} `\textit{music making bone}', aka `\textit{funny bone}' or `\textit{ulnar nerve}') rather than their technical equivalent. 

Conversely medical language might be difficult to understand for non-experts. Technical terms and a special language use make it difficult to get an easy access to information that concerns the patient. The medical science is built on a vast amount of technical expressions that are not necessarily part of a patient's everyday language. The majority of the clinical lexicon has its origin in Latin or Greek. Although the access to information is crucial for keeping track on personal conditions, for most patients the structure of the medical language remains obscure. Thus, understanding medical articles and most importantly understanding our own clinical reports written by our attending doctor may raise some challenges. In order to understand a possible serious health condition faster, automatic methods might help to simplify technical language. However, as most resources concern English language, a technical-laymen translation (and vice versa) for non-English raises further issues.

To address those challenges, this work introduces new data sets for German which support the linking of medical laymen language to technical language. Firstly we introduce a new corpus which annotates medical laymen language and technical language in a patient forum. Additionally we introduce two data sets which include different synonyms of medical concepts and sort them by complexity (rather technical to rather laymen). All data sets described in this paper will be made available\footnote{\url{http://biomedical.dfki.de}}. Our corpus in combination with the additional resources can serve as a baseline to train and to evaluate systems to map laymen into technical language and vice versa. 

\section{Related Work}
In recent years, the biomedical domain has become an important field of research for natural language processing tasks. Enhancing the patient's understanding of clinical texts is one major objective. The automatic processing of medical free text is one obstacle that is addressed by these research efforts. One step towards the processing is the mapping from free text-expressions to structured representations of domain knowledge. This includes the detection of technical terms and the normalization to an appropriate knowledge base. Synonymous expressions, terminological variants and paraphrases as well as spelling mistakes and abbreviations occur frequently in natural texts. By linking them to one unique concept, the lexical information in the text is structured and unified. In the context of medical language, different approaches face the normalization of medical concepts, such as in \newcite{leaman2013dnorm}, \newcite{suominen2013overview} or \newcite{dougan2014ncbi}.

Systems and methods that particularly address the transition from medical technical language to lay language often pursue similar approaches. Under these conditions, the linked knowledge base must provide lay language synonyms or simplified explanations for technical terms.  
In \newcite{zeng2007}, the Unified Medical Language System (UMLS) and especially the Consumer Health Vocabulary (CHV) are used as sources of lay vocabulary knowledge. \newcite{abrahamsson2014} conduct a synonym replacement for medical Swedish, using a system which assesses the difficulty of technical terms. If the technical term is considered as more difficult than the corresponding entry in the Swedish MeSH, the terms are replaced. 

 
Apart from approaches that aim at simplifying the technical language, also the mapping of laymen language to medical technical expressions has gained attraction. Social media texts are a thriving resource for genuine lay language use. Recognizing meaningful elements and linking these expressions to technical counterparts allows structured insights into the health status or health related behaviour.

For example, \newcite{oconnor2014} create a data set of annotated tweets with potential adverse drug reactions. The authors test a lexicon-based approach to detect the concepts of interest. \newcite{limsopatham2015} improve this baseline in order to normalize medical terms from social media messages using a phrase-based machine translation technique. The authors also present a system which learns the transition between lay language used in social media and the formal medical language used in descriptions of medical concepts in a standard ontology \cite{limsopatham2016}. 

Recently the Shared Task of \textit{Social Media Mining for Health (SMM4H)} has gained much interest and targets this topic as well. Some of the tasks involve for instance classification of tweets presenting adverse drug reactions or vaccine behavior mentions, see \newcite{weissenbacher2019} for more information.

Now that we introduced work related to make technical expressions more comprehensible and methods to map laymen expressions to their precise equivalent and vice versa, something still remains unclear: What actually are laymen expressions and how are medical technical expressions defined?

Previous and related work does not provide a clear definition for both. \newcite{elhadad2007} make use of the contrast between a text written by a medical professional (scientific articles) and a text written by a journalist, addressing a lay audience. They consider a term as an appropriate lay expression if it is the most frequent candidate in the lay texts. 

\newcite{chen2017} provide a method to rank medical terms extracted from electronic health records. The higher a term is ranked, the more urgently a lay translation is needed. Therefore they consider unithood, termhood, unfamiliarity and quality of compound term as relevant criteria for terms that must be translated for a lay audience. 
In contrast to these vague definitions, \newcite{grabar2014} concentrate on terms that show neoclassical compounding word formation. Consequently words with Latin or Greek roots are seen as technical terms.

\begin{table}[!h]
\centering
\begin{tabular}{p{7.5cm}}
\toprule
\textbf{Definition 1:} (a) A medical technical term is that which is used by physicians whereas (b) a medical lay term can be easily understood by patients (medical non-experts).\\
\midrule
\textbf{Definition 2:} (a) A medical term which includes (at least in parts) words with a Latin or Greek origin is defined as medical technical term. (b) All other terms belong to lay language. Lay terms are based on everyday words/language.  \\
\bottomrule
\end{tabular}%
\caption{Definitions used in this work of medical technical terms and laymen expressions}
\label{tab:Definitions}%
\end{table}%

As there is no clear definition for technical and lay expressions, we decide to incorporate the mentioned aspects and use the definitions in Table \ref{tab:Definitions}. Both definitions are not entirely satisfactorily. The first definition is subjective, depends on the background of a person and requires potentially a manually generated gold standard data set. Moreover, there might be words which belong to both groups, as they are used by physicians and at the same time are understood by patients, such as \textit{cancer}. The second definition makes it much easier to differ between both language types. However, also Latin or Greek rooted words can be very common in our daily language thus be easily understood by medical non experts, such as
\textit{hallucination}. 

\begin{table*}[!htbp]
\centering
\begin{tabular}{p{1.5cm}p{7cm}p{7cm}}
\toprule
 \textbf{Forum} & \textbf{Example} & \textbf{Translation} \\
\midrule
Stomach-Intestines & Ja. Der Termin ist tats{\"a}chlich durch.
Ich wurde an den Nieren geschallt die dort unauff{\"a}llig aussehen. (Kp was das schon ausschlie{\ss}t)
24h Urin w{\"u}rde abgegeben und eine 24h Blutdruckuntersuchung angeordnet.
Die haben mich komplett zerlegt: EKG Blut Spontanurin. &
Hi, I am very unsure at the moment, my doctors have different opinions, some doctors say that my kidneys are not looking well, the others say that I should not be worried until GFR decreases, but what is right? \\
\midrule
Kidney & Hallo, ich bin momentan sehr verunsichert, meine {\"A}rzte sind nicht gleicher Meinung, die einen {\"A}rzte sagen meine Nieren sehen nicht gut aus, die anderen sagen, solange der GFR nicht f{\"a}llt muss ich mir keine Gedanken machen, was stimmt den nun? &
Yes, the appointment is really over. The renal ultrasound showed no pathologies. (no idea what it can rule out)
I gave 24 urine sample and a 24h blood pressure test was ordered.
They have analyzed me completely: EKG, blood analysis, urine test.
\\

\bottomrule
\end{tabular}%
\caption{Excerpt of patient forum in German and (translated) English}
\label{tab:ForumExample}%
\end{table*}%

\begin{table*}[htbp]
\small
\begin{tabular}{lp{7.9cm}p{6.5cm}}
\toprule
\textbf{Tag} & \textbf{Example} & \textbf{Annotation} \\
\midrule
L & Blut im Urin (\textit{blood in urine}) & H{\"a}maturie (\textit{haematuria})\\
& Hexenschuss (lit.: \textit{a witch's shot}) & Lumbago (\textit{lumbago})\\
& Eiweissverlust {\"u}ber die Nieren (\textit{protein loss through kidneys}) & Proteinurie (\textit{proteinuria}) \\
& Durchfall (lit.: \textit{fall through}) & Diarrh{\"o} (\textit{diarrhea}) \\
& Nierenstein-Zertr{\"u}mmerung (\textit{smashing of kidney stones}) & Extrakorporale Sto{\ss}wellenlithotripsie (\textit{extracorporeal shockwave therapy})\\
\midrule
T & Aerophagie (\textit{aerophagy}) & Luftschlucken (\textit{air swallowing})\\
& Appendizitis (\textit{appendicitis}) & Blinddarmentz{\"u}ndung (\textit{appendix infection}) \\
\bottomrule
\end{tabular}%
\caption{Annotated examples of both tags (\textbf{L}ay, \textbf{T}echnical) from the Technical-Laymen Corpus, including translations}
\label{tab:Annotation Scheme Examples}%
\end{table*}%

\section{Technical-Laymen Corpus}\label{med1corpus}

This section introduces the Technical-Laymen Corpus (TLC) an annotated forum based on Med1.de\footnote{\url{https://www.med1.de/forum/}}. Med1 is a German patient forum that provides a large variety of health related topics. Users are non-professionals who seek for exchange, opinions and advice. 
Med1 is freely accessible and the discussions can be read without being registered. A registration is necessary to participate in the discussion. The operating team of Med1 does not provide medical consultation, however they guide the community in terms of netiquette. The users are anonymous and only their usernames are known to us. We would have been prepared to anonymize any personal data but we did not encounter data that could link to someone.

We are mainly interested in the medical language that is used by patients and medical laymen. A non-professional forum is likely to show the biggest source of lay language use. A corpus consisting of this kind of data should give the most realistic impression of the medical lay language. The annotation of technical and lay expressions should provide valuable insights into the relationship of technical and lay language. 


For this work we selected two subforums, namely \textit{kidney diseases} and \textit{stomach and intestines} as text source. Each subforum provides a variety of user questions (``threads''), each containing a varying number of corresponding answers. We crawled posts of the two subforums, including the time of posting, the author's nickname and the thread title. As the forum continuously grows, the corpus only represents the forum's status of the crawling date. Table \ref{tab:ForumExample} shows two exemplary sentences from the patient forum. The examples show characteristic entries in the forum, including a specific syntax and spelling errors.


\subsection{Annotation Schema}
Mainly we are interested in terms and expressions that are used by medical non-professionals as those provide a large variety which cannot be entirely covered in medical dictionaries. However, as people might undergo a lifelong treatment (kidney diseases are chronic diseases) patients are well informed and also use frequently technical terms and abbreviations. For a newbie this might be difficult to understand. Thus, we target also the other direction -- the detection of technical terms in order to simplify them. Our annotation involves two different concepts: (1) \textit{lay expressions} and (2) \textit{technical expressions}. Regarding those information we mainly focus on \textit{symptoms}, \textit{diseases}, as well as \textit{treatments and examinations}.
However annotators were free to also label information that goes beyond the focus information (e.g. body parts, medication).

Annotators were asked in case of a lay expression to include the corresponding technical counterpart as well, and in case of a technical expression, the most common lay expression. We opt for a single word counterpart. If this is not possible, we choose a paraphrase or a short, appropriate explanation.
In case of abbreviations we treat them accordingly: If the abbreviation is presumably known to a layman or even typical layman use (e.g. \textit{KKH} for ``Krankenhaus", \textit{hospital}), we annotate it as a lay expression. If the abbreviation is untypical or unlikely to be known to a patient (e.g. \textit{NBE} for ``Nierenbeckenentz{\"u}ndung", \textit{Inflammation of the Renal pelvis}) we treat it as technical term. In both cases we add the expanded version.
Table \ref{tab:Annotation Scheme Examples} presents examples of the categories including their English translation.

\subsection{Annotation Setup and Process}

The annotation has been then carried out by two medical students within various iterations using the brat \footnote{\url{http://brat.nlplab.org/}} annotator tool \cite{stenetorp:2012}. The first annotation cycle concentrated on medically obvious cases. This means that we focused on medically clear translations from lay to technical language or vice versa. For example, the term ``Normotonie" (\textit{normotonia}) is assigned the tag \textit{technical} and the corresponding lay expression ``normaler Blutdruck" (\textit{normal blood pressure}) is given as free text.

However the results of the first cycle were not satisfying yet, as most translations were already well documented in existing vocabularies. Therefore we extended the annotations by including cases in which a non-professional describes a medical concept in such way that a definite technical translation is difficult. For example, if a user describes \textit{problems with passing water} (``Probleme beim Wasserlassen"), a possible technical equivalent could be \textit{dysuria}. 

From the medical point of view, this procedure is difficult because it includes to some extend interpretation work: While \textit{problems with passing water} is only a rough symptom description, a dysuria is a pathological state. The transfer from a symptom description to a disease can be seen as kind of diagnostic process which must be avoided at that point. As the annotation was carried out by medical students we trusted their expertise to decide at which point the annotation would exceed a reasonable interpretation. Thus we do not opt for a diagnostic interpretation of symptoms. In order to retrace such cases, the annotators highlighted annotated terms that came close to a critical interpretation level.  

Within a final iteration one of the authors examined the annotations and highlighted potential errors (wrong labels, missing information etc.). Those highlighted information were then again manually examined, in order to provide a corpus with an appropriate quality.

\subsection{Corpus Analysis}

Table \ref{tab:Med1 Corpus} provides an overview about TLC. The table lists for each forum topic the number of included files, number of tokens, as well as the average number of tokens per file and the average number of annotations per file. Note that not all files included relevant information to be annotated. A more detailed overview about the annotated information itself is presented in Table \ref{tab:Annotation Scheme}. The table lists the the number of overall and number of unique annotations for each label. As the table shows, the most annotated labels are laymen expressions. Moreover those expressions also have the largest variety in terms of different unique terms. This makes sense and highlights the importance detecting laymen expressions. 

\begin{table}[htbp]
\centering
\small
\begin{tabular}{lcc}
\toprule
& \textbf{Kidney} & \textbf{Stomach-Intestines} \\
\midrule
Number of files & 2000 & 2000  \\
Number of tokens & 203,553 & 234,914 \\
Avg. tokens /file & 101.78 & 117.46 \\
Avg. annotations /file & 2.52 & 1.41 \\
\bottomrule
\end{tabular}%
\caption{General overview about Med1 Corpus}
\label{tab:Med1 Corpus}%
\end{table}%

\begin{table}[htbp]
\centering
\small
\begin{tabular}{rcc}
\toprule
\textbf{Label} & \textbf{\#Annotations} & \textbf{\#Unique}\\
\midrule
Lay Expression  & 4727 & 1246\\
Technical Term & 1745 & 376\\
\bottomrule
\end{tabular}%
\caption{Overview about number of annotated and unique concepts of each category label.}
\label{tab:Annotation Scheme}%
\end{table}%

\begin{table*}[!htbp]
\centering
\small
\begin{tabular}{p{2.3cm}p{7.5cm}p{5cm} }
\toprule
\textbf{Term} & \textbf{Explanation} & \textbf{Synonym}  \\
\midrule
Dialyse & Anwendung der Dialyse, vor allem zur Reinigung von Blut & Blutreinigung; Blutw{\"a}sche \\
Diabetes & Stoffwechselerkrankung, bei der eine gesteigerte Unempfindlichkeit gegen{\"u}ber Insulin besteht (sogenannter Diabetes mellitus Typ 2 oder Typ-2-Diabetes oder Altersdiabetes) & Zuckerkrankheit; Zucker \\
Delirium tremens & Ernste und potentiell lebensbedrohende Komplikation im Alkoholentzug bei einer schon l{\"a}nger bestehenden Alkoholkrankheit & Alkoholdelir; {\"O}nomanie; S{\"a}uferwahn; S{\"a}uferwahnsinn \\

\bottomrule
\end{tabular}%
\caption{Example of extracted information from Wiktionary}
\label{tab:Wiktionary}%
\end{table*}%

\begin{table*}[!ht]
 \centering
 \small
 \begin{tabular}{ lllllll }
   \toprule
   \textbf{CUI} & \textbf{English} & \textbf{German} & \textbf{Spanish} & \textbf{French} & \textbf{Swedish}  & \textbf{Russian} \\
   \midrule
   C0007097 & carcinoma & Karzinom & carcinoma & carcinome & Karcinom  & KARTSINOMA \\
  C0012503 & Dioxins & Dioxine & Dioxinas & Dioxines & Dioxiner & DIOKSINY  \\
  C0023531 & Leukoplakia & Leukoplakie & Leucoplaquia & Leucoplasie & Leukoplaki & LEUKOPLAKIJA  \\
  C0027804 & Neurasthenia & Neurasthenie & neurastenia & Neurasth\'enie &  Neurasteni  & NEVRASTENIIA  \\
  \bottomrule
\end{tabular}
\caption{Similar mentions of different languages in UMLS linked by the same concept unique identifier (CUI).}
\label{Tab:similarityAcrossLanguages}
\end{table*}




\section{Additional Resources and Methods to Process Technical-Laymen Language}\label{resources}

In addition to the Technical-Laymen Corpus we extract data from two additional resources: UMLS and Wiktionary. We aim at providing assorted data sets which incorporate a matching of technical and laymen language in the biomedical domain. 
Both resources are processed and can be used to support the linking from laymen to technical terms and vice versa. However as both resources do not systematically differ between lay and technical terms, we additionally propose a simple method to identify technical (and less technical) terms.


\subsection{UMLS Synonym Subset}

The Unified Medical Language System (UMLS) is a biomedical ontology and knowledge source. The Metathesaurus of UMLS provides a vocabulary database for the biomedical and health domain. Synonymous expressions are linked by the same concept unique identifier (CUI). The same CUI also links equivalent expressions in different languages. The Semantic Network of UMLS categorises all terms into broad subject categories, providing a categorization into 127 semantic types (STY) and 54 relation types (RL).
Overall UMLS includes concepts of over 34 million concepts in English language, whereas only approximately 100,000 in German. Roughly half of those concepts include at least two mentions. While the German UMLS subset is relevant for concept normalization in general, particularly concepts including synonyms are interesting, as they might include technical and laymen expressions. 

\subsection{Wiktionary Synonym Subset}

Our second resource is build from the German version of Wiktionary\footnote{\url{https://de.wiktionary.org/wiki/Wiktionary:Hauptseite}}. Wiktionary provides 741,260 (Jan 2019) entries in German. Although biomedical information is not a special focus of Wiktionary, there is a large range of related subcategories. In order to create our technical/laymen language resource the (in November 2019 newest) German Wiktionary dumb has been downloaded and further processed and filtered to our needs. 
In order to build a technical/laymen language resource from Wiktionary, we parsed the provided dump and automatically gathered for each entry the term, its explanation and, if available, synonyms. Our focus is the biomedical domain, thus we limited the data by selecting medical related entries only. These entries come from the categories \textit{Medicine, Pharmacy, Pharmacology, Anatomy, Psychiatry, Psychology, Physiology, Ophthalmology, Pathology, Dentistry, Gynaecology} and \textit{Dermatology}. Additionally, we included every entry that contains at any place the regular expression \textit{krank} (\textit{sick}) which should relate to mentions of diseases. By doing so, the resulting resource is larger than necessary (e.g. some veterinary entries are included). However we ensure to make use of all entries that could be relevant. Only entries of the mentioned categories were used for our resource.
The final biomedical Wiktionary subset comprises 4468 concepts and nearly all including a definition. 2155 of the entries include at least one synonym. Overall this subset includes 8657 different entries.

Even though the data set appears to be small in comparison to UMLS, an interesting aspect about Wiktionary is the variety of laymen synonyms. It includes lay expressions which are often not covered by UMLS. Table \ref{tab:Wiktionary} shows some examples: Diabetes for instance is a characterized by recurrent or persistent high blood sugar. A non-professional German term for diabetes is ``Zuckerkrankheit" (lit.: \textit{sugar disease}) or simply ``Zucker" (\textit{sugar}). These terms, even though frequently used, are not listed in UMLS. 
The large variety of lay expressions includes not only lay expressions to the respective technical term but also colloquial or even vulgar terms. For example, the entry of ``Diarrhoe" (\textit{diarrhea}) lists as synonyms ``Schnelle Katharina" (\textit{fast Katharina}) and ``Flotter Otto" (\textit{quick Otto}).


\subsection{Aligning data sets}

UMLS is frequently used for concept normalization and it comprises much more concepts than the Wiktionary subset. Conversely, Wiktionary appears to be a highly useful resource as it contains more casual expressions in medical context. For this reason we try to combine both data sets. For this, we identify expressions from Wiktionary which also occur in UMLS. If a term from Wiktionary also occurs within exactly one CUI in UMLS, we can simply align the Wiktionary concept with all its synonyms to this CUI. For instance if the Wikitonary term `pain' (and all its synonyms) would occur only in context of one single UMLS-CUI, we can map the Wiktionary term `pain' and all its synonyms to this corresponding CUI. However, this is not possible in all cases, as terms in UMLS might be assigned to various CUIs.

In this way, 768 CUIs can be extended by overall 3082 additional mentions. We refer to the resulting data set as Wiktionary-UMLS (WUMLS).

\begin{table*}[!htbp]
\centering
\small
\begin{tabular}{lccccccccccc}
\toprule

distance ($>$=) & 0 & 5 & 10 & 15 & 20 & 25 & 30 & 35 & 40 & 45 & 50 \\
\midrule
\#instance & 300 & 237 & 193 & 161 & 144 & 124 & 97 & 87 & 74 & 56 & 49 \\
\%is-easier & 50 & 59 & 65 & 71 & 74 & 74 & 75 & 74 & 70 & 70 & 71 \\
\%is-easier-or-equal & 88 & 89 & 91 & 92 & 92 & 93 & 93 & 92 & 91 & 89 & 88 \\

\bottomrule
\end{tabular}
\caption{Manual Evaluation of 300 selected examples to explore if the term ranked as easiest term is in fact easier than the term ranked as most technical. Considering only pairs with a larger edit distance, the results show that precision increases for both \textit{is-easier} (checking whether the term is in fact simplified) or \textit{is-easier-or-equal} (checking whether the term is at least not more complicated).}
\label{tab:UMLS_Eval}
\end{table*}

\subsection{Sorting Synonyms}

The mapping from technical to laymen language is one of the aspects of this work. However, the largest of our supporting resources, UMLS, does not provide any information about technical or laymen language for German. For this reason we provide a simple technique to identify technical and less technical terms according to definition 2 (see Table \ref{tab:Definitions}). According to this, technical terms have their origin in Latin or Greek language. Moreover, we know that those technical terms are very common in many (particular European) languages. Table \ref{Tab:similarityAcrossLanguages} shows examples of similar expressions across various languages. Using this characteristic we propose the following method to identify medical technical expressions:




For each German target mention ($G_{t}$) we identify the English ($E_{j}$) and French ($F_{k}$) synonym with the lowest Levenshtein distance ($lev(a,b)$) for each of both languages. Next we calculate the average between both minimum distance scores. Note, we chose two languages rather than one to have a more robust distance score. Finally we harmonize this score, dividing it by the length of the target mention ($len(a)$). This should avoid that short strings are favoured over longer strings with similar edits. We refer to this score as the \textit{harmonized distance} (\textit{h\_dist}). The harmonized distance can be formulated as follows:

\begin{equation}
h\_dist_{G_{t}} = \frac{min(lev(G_{t},E_{j}))+min(lev(G_{t},F_{k}))}{2*len(G_{t})}
\end{equation}


\paragraph{Sorted Synonym data set (SSD):}\label{sorted_syn} Following the assumption from above, we assume that a German mention with a low \textit{harmonized distance} might likely to have a Greek or Latin origin, thus tends to be a technical term. Thus we calculate the \textit{harmonized distance} of all German mentions of UMLS (and WUMLS) and sort all synonyms of each concept according to this score. Starting with the term with the lowest distance score and finishing with the one with the largest score. 

As we are interested in particular concepts we select only those which belong to one of these semantic types (STY): `\textit{Anatomical Abnormality}', `\textit{Anatomical Structure}', `\textit{Body Location or Region}', `\textit{Body Part, Organ, or Organ Component}', `\textit{Body Space or Junction}', `\textit{Disease or Syndrome}', `\textit{Injury or Poisoning}', `\textit{Mental or Behavioral Dysfunction}', `\textit{Sign or Symptom}'. Using the technique from above and including English and French as reference language, we can generate sorted synonym sequences of 28,495 different concepts with overall 47,996 different mentions.

\paragraph{Evaluation 1 -- Are synonyms with a low harmonic distance technical terms?} 
In order to examine this question we randomly select 300 concepts and their lowest $h\_dist$ mention from UMLS-SSD. All selected mentions had a different harmonic score, whereas the largest score of the subset was 120. The selected mentions have been manually evaluated according to our two definitions by one of the authors. The analysis shows that 75\% of all terms are technical expressions according to definition 1 and 90\% according to definition 2. Table \ref{tab:UMLS_Eval1} shows an analysis considering only concepts below a certain harmonic distance threshold. In this way we can see that a harmonic distance below 60 leads to a high accuracy, which supports our assumption. The larger the distance the more the accuracy decreases. However the score decreases faster using definition 1.

\begin{table}[htbp]
\centering
\small
\begin{tabular}{lcccccc}
\toprule

distance ($<$=) & 20 & 40 & 60 & 80 & 100 & 120 \\
\midrule
\#instances & 59 & 105 & 174 & 277 & 297 & 299 \\
\%definition-1 & 93 & 93 & 91 & 79 & 75 & 75 \\
\%definition-2 & 98 & 99 & 99 & 94 & 90 & 90 \\

\bottomrule
\end{tabular}%
\caption{Manual examination of 300 randomly selected expressions of a concept with the lowest harmonic distance score.}
\label{tab:UMLS_Eval1}%
\end{table}%

\paragraph{Evaluation 2 -- Are synonym mentions with a larger harmonic distance less technical and possibly laymen expressions?} 
In order to examine this question we examine whether the term with the lowest score in UMLS-SSD is more or at least similarly technical as the term with the largest score of all synonyms. Thus, we selected randomly 300 German concept mention pairs, this time with the lowest and the largest \textit{harmonic distance score} and examined whether the first term is a) more technical, b) similar technical or c) less technical than the second term. As we do not know whether there is always a simplified term within the synonym set, we evaluate according to \textit{is-easier} ($a/(a+b+c)$), as well as \textit{is-easier-or-equal} ($(a+b)/(a+b+c)$). 

The results in Table \ref{tab:UMLS_Eval} show that in only 50\% of the cases the expression with the highest \textit{harmonic distance} is less technical than the expression with the lowest \textit{harmonic distance}. This does not look very promising at first. However we can make the following analyses: First considering all synonym pairs, in 88\% of the cases the expression with the highest \textit{harmonic distance} is easier or at least similarly technical as the expression with the lowest score. Moreover the table shows that the absolute distance between both scores has a strong influence on the outcome. Increasing the absolute distance between both scores quickly increases also the accuracy (\%). In case of examining whether the expression with the higher score is in fact less technical, we can see a constant increase from 50\%, using all pairs, to 75\% considering a minimum absolute distance of 30. Increasing the distance, decreases obviously the number of synonym pairs. However, after reaching a maximum of 75\%, the scores drop slightly, but never undergo 70. A similar effect can be observed for \textit{is-easier-or-equal}. After a maximum of 93\% with a distance of 30, the values slightly decrease but remain always above 88. 

Overall these results are very promising. Considering a certain distance (e.g. of 15 or more), we can ensure that in more than 70\% of the cases the synonym with the larger harmonic distance is less technical and in 92\% of the cases the term is at least not more complicated.


\section{Baseline Experiments}\label{experiment}

In the previous sections we presented the TLC corpus and in addition two further resources to support the mapping between German medical laymen to technical language and vice versa. The main focus of our work is the presentation of new resources in this domain. In this section, however, we present in addition some baseline results on TLC which can be used as benchmark for future work.

Regarding baseline results, we carry out two different experiments: 1) the normalization of medical technical terms including a term simplification and the 2) normalization of medical laymen expressions. For our experiment we indexed the mentions (and its stemmed version) from UMLS/WUMLS in Solr.

\subsection{Experiment 1 -- Normalization and Simplification}

For experiment 1 we extract all technical terms 
and examine whether we can align it to a corresponding concept unique identifier. Using UMLS in 72.10\% of the cases we can find the corresponding medical concept. However only in 31.11\% of those cases we find an easier synonym. The usage of WUMLS does not increase the performance much. However if we analyse the terms found in UMLS in more detail, we can see that the average harmonic distance score of those expressions is 39.93. As we know from Evaluation 1 in Section \ref{sorted_syn} that a low score is an indicator for a technical term, this score is no surprise. We can also see that a large number of expressions include a larger harmonic distance, for instance 143 expressions have a score of 70 or above. 


\subsection{Experiment 2 -- Normalization Laymen Language}

For experiment 2 we extract all laymen terms and examine whether the corresponding technical term can be found. In case of using UMLS terms for only 57.37\% of the mentions a corresponding CUI can be detected. As laymen expressions provide much more variations in comparison to technical terms, this outcome was expected. If we again examine the expressions found in UMLS in more detail we can see that the average \textit{harmonic distance} is at 82.05. However also here we can find a large number of expressions supposed to be non-technical, but have a low \textit{harmonic distance}. For instance 137 expressions have a score below 170.

Finally, using WUMLS data for the normalization the score can be increased to 64.08\%. This shows clearly the advantage of including additional information of Wiktionary.

\subsection{Discussion}

Overall the results of our baseline experiments show that laymen language concept normalization is much more difficult in comparison to the normalization of medical technical expressions. This highlights the importance of creating further resources of laymen synonyms but also methods being able to map between those language types. 

Methods trained on definitions such as in \newcite{limsopatham2016} might be helpful to tackle this challenge. However, in comparison to English UMLS and also Wiktionary do not contain as many German definitions as for English language. This again highlights the aspect that German, in comparison to English, is a low resourced language considering existing and freely available structured resources. As mentioned above, the German UMLS subset covers only 3.2\% of all English concepts and involves only 2.3\% of all existing English synonyms. Thus, it is obvious that concept normalization even for technical terms is much more challenging. Cross-lingual methods such as in \newcite{roller2018b} might help to increase the coverage of technical terms.


\section{Conclusion}
In this work we presented a new corpus based upon a patient forum for kidney disease and stomach-intestines. The data set labels medical laymen language and technical terms and assigns a corresponding description or expression. This resource might be valuable resource to map and translate between both types of language styles in the medical domain. In addition to that we also provided two resources which can support this translation process. Finally we also tested a simple baseline on our corpus which can be used as reference for more complex methods.

\section*{Acknowledgments}
This project was funded by the European Union's Horizon 2020 research and innovation program under grant agreement No 780495 (BigMedilytics) and by the German Federal Ministry of Economics and Energy through the project MACSS (01MD16011F).

\section{Bibliographical References}
\label{main:ref}

\bibliographystyle{lrec}
\bibliography{lrec}


\end{document}